\documentclass{IOS-Book-Article}
\usepackage{subcaption}  

\usepackage{amsmath}
\usepackage{graphicx}
\usepackage{soul}\setuldepth{article}
\usepackage{xcolor}
\usepackage{multirow}
\usepackage{xspace}
\newcommand{\mstd}[2]{\ensuremath{#1{\scriptstyle \pm #2}}}
\usepackage{colortbl}
\usepackage{booktabs} 
\usepackage{array}   
\usepackage{amssymb}
\usepackage{bbm}

\newcommand{\TReNN}{\textsc{TReNN}\xspace}
\newcommand{\ReNN}{\textsc{ReNN}\xspace}
\newcommand{\TNN}{\textsc{TNN}\xspace}
\newcommand{\SNN}{\textsc{SNN}\xspace}
\newcommand{\TEReNN}{\textsc{TE-ReNN}\xspace}

\usepackage{wrapfig}

\usepackage{pifont}
\usepackage{xcolor}
\newcommand{\cmark}{\textcolor{green!50!black}{\ding{51}}} 
\newcommand{\xmark}{\textcolor{red!70!black}{\ding{55}}}   

\def\hb{\hbox to 11.5 cm{}}

\begin{document}

\pagestyle{headings}
\def\thepage{}
\begin{frontmatter}              

\title{Actionable Real-Time Modeling of Surgical Team Dynamics via Time-Expanded Interaction Graphs}

\author[A]{\fnms{Vincenzo Marco} \snm{De Luca}}
\author[B]{\fnms{Antonio} \snm{Longa}}
\author[A]{\fnms{Giovanna} \snm{Varni}}
\author[A]{\fnms{Andrea} \snm{Passerini}}
\address[A]{University of Trento}
\address[B]{UiT the Arctic University of Norway}

\begin{abstract}
Surgical team performance arises from complex interactions between technical execution and non-technical skills, including communication and coordination dynamics. However, current surgical AI systems predominantly model visual workflow signals, lacking structured representations of intraoperative team interactions over time. We propose a real-time actionable approach for modeling surgical team dynamics using time-expanded interaction graphs, where team members are modeled as time-indexed nodes and communication exchanges define directed edges. This spatio-temporal expansion enables dynamic interaction modeling, while allowing efficient inference with a static graph neural network. 
The model predicts procedural efficiency as the deviation from the expected 
duration 
and supports real-time deployment. Beyond prediction, we perform a counterfactual analysis to identify minimal changes in communication structure and interpretable behavioral variables associated with improved predicted outcomes. 
Experiments on recorded surgical procedures show that structured modeling of team interactions improves early identification of prolonged interventions and provides coherent, actionable explanations. This work advances surgical AI toward real-time, team-aware, and actionable decision support in the operating room.
\end{abstract}

\end{frontmatter}

\section{Introduction}

Actionable AI refers to systems that go beyond prediction or explanation 
by providing recommendations that users can act upon to achieve desired outcomes while considering relevant trade-offs~\cite{shrestha2019organizational}.
High-stakes scenarios particularly benefit from Actionable AI, where predictive accuracy alone is insufficient and decision support systems must translate model outputs into actionable feedback.
The operating room (OR) exemplifies these settings due to its intrinsic complexity and the critical role played by teamwork in ensuring safe and efficient procedures~\cite{smit2023future}.
 Fields such as Surgical Data Science (SDS) aim to improve interventional healthcare through procedural data analysis~\cite{vedula2017surgical, ward2021surgical}. However, current SDS research predominantly focuses on visual workflow modeling~\cite{garrow2021machine, liu2025deep} and technical skill assessment~\cite{lam2022machine,kostopoulos2025prediction}, while intraoperative team dynamics remain largely unexplored~\cite{lalys2014surgical, harari2024deep, mendu2025voice}, despite team performance emerging from complex interaction processes~\cite{etherington2021measuring, watkins2023team}. 
Moreover, existing studies that consider team dynamics are typically limited to predicting social signals, without providing actionable guidance to support performance improvement. 
Communication breakdowns, misaligned task allocation, and micro-coordination failures may accumulate and introduce temporal inefficiencies~\cite{torring2019communication} that translate into increased postoperative risk~\cite{cheng2018prolonged}. A recent study on spinal surgery shows that 15\% increase in operative time is associated with a 14\% higher risk of complications~\cite{monetta2024prolonged}.
Although not a direct measure of care quality, systematic deviations from the expected operative time provide a practical proxy for team performance~\cite{stucky2024surgical}, with prolonged surgeries impacting both patient safety and costs, up to +160\% per procedure~\cite{roach2022cost}, and incurring substantial OR expenses, up to \$37 per minute~\cite{childers2018understanding}.


In this work, 
we frame operative time as a predictive and actionable signal of team performance. We model team interactions by jointly capturing temporal and relational dimensions from multimodal communication data. While temporal graph neural networks have demonstrated strong performance in modeling spatio-temporal data~\cite{longa2023graph}, they are less suitable in small-data regimes such as surgical teams where the limited amount of data and the team size increase overfitting risk~\cite{pmlrv269luca25a}. To address this challenge, we incorporate temporal dynamics directly into relational networks through the construction of time-expanded graphs. Finally, we introduce a counterfactual analysis framework to identify interpretable behavioral changes in team interactions associated with improved predicted outcomes.
The main contributions of this paper can be summarized as follows:




\begin{itemize}
\item A Time-Expanded Relational Neural Network (\TEReNN) architecture that simultaneously models relations and temporal evolution for teams.
\item A 
two-level counterfactual algorithm to generate behavioral suggestions to enhance team performance.
\item An experimental evaluation on simulated surgical procedures showcasing the effectiveness of the proposed solution in both improving early identification of prolonged
interventions and providing actionable counterfactuals. 
\end{itemize}

\begin{figure}[t!]
    \centering
    \includegraphics[width=1\linewidth]{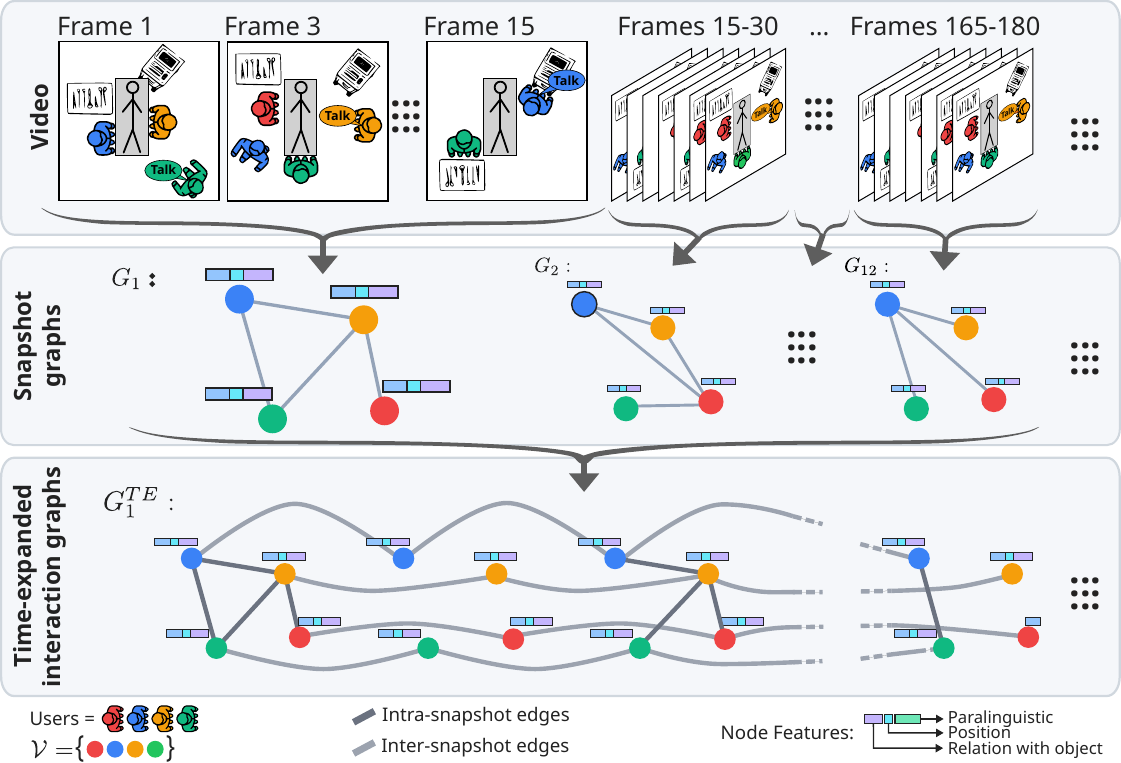}
    \caption{Overview of the interaction modeling pipeline. 
    Top row: multimodal time-series are segmented into fixed temporal windows (15 seconds). 
    Middle row: for each window, a snapshot interaction graph $G_t$ is built, in which nodes represent team members and edges encode broadcast verbal communication. Node features integrate interpretable paralinguistic (eGeMAPS), pose, and human–tool interaction. 
    Bottom row: snapshot graphs are connected through identity-preserving temporal edges into a Time-Expanded Interaction Graph $G^{TE}$, enabling joint modeling of intra-team coordination and longitudinal individual dynamics across time.}
    \label{fig:mainpanel}
\end{figure}

\section{Methodology}
Figure~\ref{fig:mainpanel} illustrates the proposed pipeline of our approach, from multimodal feature extraction to the construction of the Time-Expanded Interaction Graph. 
The methodology is designed not only to capture team interaction dynamics but also to preserve an explicit link between model representations and modifiable human behaviors, a key requirement for Hybrid Human-AI collaboration.

\subsection{Interaction Modeling Framework}

We model surgical team collaboration as a sequence of multimodal interaction graphs constructed over fixed temporal windows. The input consists of paralinguistic, motion, and human-object interaction time-series features extracted from audio-video recordings and annotations. Each modality is sampled at its own \emph{ad hoc} rate.
Time windows of 15 seconds are constructed from these time series.

Let $\mathcal{V}$ denote the set of team members in the OR. 
For each temporal window $t$, we build a \textit{snapshot interaction graph} $G_t = (V_t, E_t)$,
where $V_t \subseteq \mathcal{V}$ is the set of team members present in the OR during window $t$. 
Edges $E_t$ encode verbal communication under a broadcast assumption. 
When a team member $v_i \in V_t$ speaks during window $t$, we add edges $(v_i, v_j)$ for all $v_j \in V_t$, $j \neq i$. 
This representation models verbal communication as being directed to all present team members as a one-to-many interaction, rather than to a specific addressee. 
While simplified, this assumption reflects the shared acoustic environment of the OR, where speech is typically audible to all team members. Each node $v \in V_t$ is associated with a multimodal feature vector integrating:

\begin{itemize}
    \item \textbf{Paralinguistic features}: loudness, alpha-ratio, and harmonics-to-noise ratio features from the eGeMAPS 2.0 set~\cite{eyben2015geneva}.
    \item \textbf{Motion features}: the position of each team member, and the average displacement and its standard deviation.
    \item \textbf{Human-tool interaction features}: a triplet of values (team member-action-object) describing atomic actions in the OR (e.g., anesthesiologist-calibrating-instrument). 
    These features are one-hot encoded.
    \item \textbf{Role features}: the functional role of each team member 
    (e.g., head surgeon, nurse). Roles' features are one-hot encoded.
\end{itemize}


\subsection{Behavioral State Representation}
\label{behavioral_state_representation}

To support interpretable and actionable suggestions in the form of counterfactual explanations, we focus on a subset of paralinguistic features whose meanings directly map to interpretable behavioral patterns. As previously mentioned, the paralinguistic features are the loudness, referring to the activation and engagement, the harmonics-to-noise ratio that measures vocal control and tension, and the alpha ratio accounting for the vocal dominance. Each combination of these three features is mapped to a discrete behavioral class.


\begin{table}[h!]
\centering
\scriptsize
\begin{tabular}{c c c l}
\hline
\textbf{Activation} & \textbf{Control} & \textbf{Dominance} & \textbf{Behavioral Class} \\
\hline
Absent &  Absent & Absent & Silent$^{*}$ \\
Low  & Low  & Low  & Withdrawn / Disengaged \\
Low  & Low  & High & Restrained Conflict \\
Low  & High & Low  & Calm--Cooperative \\
Low  & High & High & Quiet Authority \\
High & Low  & Low  & Agitated / Over-aroused \\
High & Low  & High & Dominant--Aggressive \\
High & High & Low  & Engaged--Cooperative \\
High & High & High & Calm Leader \\
\hline
\end{tabular}
\caption{Behavioral classes defined by combining  paralinguistic features from the eGeMAPS set.}
\label{tab:behavioral_classes}
\end{table}

The dichotomization of the paralinguistic features into High/Low categories is used exclusively for interpretability and behavioral abstraction. During model training, we keep the original continuous feature values extracted from eGeMAPS to preserve fine-grained acoustic information and predictive fidelity. 
This separation preserves interpretability without compromising modeling capacity.

\subsection{Time-Expanded Relational Neural Networks (\TEReNN)} 

To capture team coordination dynamics,  
we construct a Time-Expanded Interaction Graph, modeled through Time-Expanded Relational Neural Networks (\TEReNN), that embeds temporal evolution directly into the graph topology as: $G^{TE} = (V^{TE}, E^{TE}), V^{TE} = \{ v_i^t \mid v_i \in \mathcal{V},\; t = 1, \dots, T \}.$
Each node $v_i^t$ represents a team member $v_i$ at a specific temporal window $t$. The full graph $G^{TE}$ therefore contains multiple temporal instances of each team member, enabling the modeling of their behavior throughout the entire procedure.
To capture both instantaneous and longitudinal relationships, the edge set is composed of two complementary components $E^{snap}$ and $E^{temp}$ such that $E^{TE} = E^{snap} \cup E^{temp}$: 

\begin{itemize}
    \item \textbf{Intra-snapshot edges}: 
     $E^{snap} = \{ (v_i^t, v_j^t) \mid (v_i, v_j) \in E_t \}$.
    They capture broadcast communication from a team member $v_{i}^{t}$ to all the other team members $v_{j}^{t}$ within the same temporal window $t$.
   
    \item \textbf{Inter-snapshot identity edges}: $E^{temp} = \{ (v_i^t, v_i^{t+1}) \mid v_i \in \mathcal{V},\; t = 1, \dots, T-1 \}$.
    They link consecutive temporal instances of the same team member $(v_i^t, v_i^{t+1})$ to preserve identity continuity over time.
   
\end{itemize}
This construction encodes temporal dynamics directly into the graph structure rather than recurrent architectures. 
Information can therefore propagate both across team members within a window and longitudinally along each individual trajectory.
In our implementation, each $G^{TE}$ spans 3 minutes of audio-video recordings (i.e., 12 temporal windows), providing sufficient temporal context to capture evolving coordination patterns while remaining computationally tractable.
By combining relational structure with interpretable behavioral abstraction, the proposed representation enables the learning of predictive signals of procedural efficiency while maintaining a direct link between model representations and human-adjustable behaviors.

\subsection{Counterfactual Explanations}
\label{sec:counterfactual_xai}
In high-stakes scenarios, actionable feedback is crucial for understanding the factors underlying team performance.
We define feedback as counterfactual explanations, which provide guidance on how team behavior can be modified. Counterfactuals identify minimal changes that lead to more desirable outcomes at two different levels:  topological, indicating which interactions should be added or removed, and feature-level, indicating how a team member’s behavior should change.

The topological counterfactual explanations build on a combinatorial optimization procedure that evaluates whether adding or removing interactions results in a change of the predicted operative time. In our setting, using a speaker-broadcast strategy to define the interaction topology, we consider a team of $N$ members over $T$ time steps. We analyze $N \times T$ counterfactual scenarios by either adding or removing all interactions associated with a team member at each time step, while appropriately padding or replacing their paralinguistic features to reflect whether they were speaking or not.

Feature-level counterfactuals focus exclusively on the eight behavioral classes reported in Table \ref{tab:behavioral_classes}, corresponding to distinct interaction styles 
A counterfactual at this level corresponds to switching a member’s behavioral class throughout the entire temporal window, using the class centroid as the reference. Such a switch propagates through the team via the interaction structure, potentially improving or worsening predicted performance.

\section{Experimental setting}

Our experimental evaluation aims at answering the following research questions:
\begin{itemize}
    \item \textbf{RQ1.} Can structured modeling of team interaction dynamics improve forecasting of procedural efficiency compared to baselines?
    \item \textbf{RQ2.} Can minimal, interpretable behavioral adjustments in team communication be identified that are associated with improved predicted efficiency?
\end{itemize}

\subsection{Dataset}
The MM-OR~\cite{ozsoy2025mm} dataset is a multimodal corpus of simulated knee-replacement surgical procedures. The surgical teams consist of either expert surgeons or trainees. In total, the dataset comprises 27 surgical procedures, of which 14 correspond to complete knee-replacement surgeries. Each procedure typically involves four to six team members, with their functional roles explicitly annotated.
The OR is recorded using five cameras, while audio is captured via a single environmental microphone. The dataset also includes annotations about the use of tools from team members. 

\subsection{Pre-processing}
Automated diarization and transcription of audio recordings are carried out using Pyannote~\cite{bredin2020pyannote} and Whisper~\cite{radford2023robust}, respectively. 
However, due to the single-channel recording, ambient scene noise, and frequent language switches between English and German, such procedures are often unreliable. 
Thus, two human annotators manually corrected the automatic diarization and transcription outcomes. 
The clustering of surgeries into \emph{slow}, \emph{medium}, and \emph{fast} operative time is computed according to the mean and standard deviation of the available surgical procedures. This results in a highly imbalanced class distribution: two slow, ten medium, and two fast, in such a way to properly point out significant changes in procedural duration.
The data split is constructed to ensure that both slow and fast procedures are represented across folds, while strictly maintaining separation between training and test sets under a leave-one-team-out protocol.

\subsection{Baselines}

Multiple computational approaches have been proposed to model team behaviors, differing in their ability to capture temporal and relational dependencies~\cite{de2025boosting}. Static Neural Networks (\SNN) model individual features only, Temporal Neural Networks (\TNN) capture temporal evolution, Relational Neural Networks (\ReNN) focus on interactions, and Tempo-Relational Neural Networks (\TReNN) combine both. Our method extends these baselines by introducing \TEReNN to address the limitations of modeling small graphs by expanding the graph through its evolution over time. This approach incorporates temporal evolution into the graph topology, enabling joint modeling of intra-team interactions and longitudinal individual behavior.


\section{Preliminary results}

\subsection{Q1: Predictive performance} 
Table~\ref{tab:placeholder} summarizes a systematic comparison across progressively richer modeling assumptions, isolating the contribution of behavioral features, temporal dependencies, and relational graph structure.
We evaluate all architectures using macro-F1, due to label imbalance, under a leave-one-team-out evaluation protocol.
The values are averaged over 10 seeds.
The comparison isolates three complementary sources of information: multimodal behavioral features, temporal modeling, and relational graph topology. Results show that models relying on any single component consistently perform worse than those combining them, suggesting that these information sources provide complementary contributions to performance.
\SNN relying only on features, such as Multi-Layer Perceptrons (MLP) and Random Forest (RF), underperform compared to \TNN that explicitly captures temporal dynamics, including Long Short-Term Memory networks (LSTM)~\cite{schmidhuber1997long} and Multi-Head Attention (MHA)~\cite{vaswani2017attention}, or \ReNN that explicitly models relational structure, for instance Graph Convolutional Networks (GCN)~\cite{kipf2016semi} and Graph Attention Networks (GAT)~\cite{velickovic2017graph}.
Combining temporal and relational modeling in \TReNN (MHA+GCN, MHA+GAT) further improves performance, highlighting the complementary nature of these signals.
However, architectures that deal with temporal and relational information as separate modules remain consistently outperformed by the proposed \TEReNN formulation.
By embedding temporal continuity directly into the graph topology, our method jointly models intra-team coordination and longitudinal individual dynamics within a unified relational structure, achieving the strongest performance among all evaluated configurations.

\begin{table}[t]
    \centering
    \resizebox{\textwidth}{!}{%
    \begin{tabular}{l | l | c c c | c}
    \toprule
        Paradigm & Model & Feature & Time  & Graph topology & F1-macro (\%)\\
        \midrule
        \multirow{2}{*}{\centering \SNN} 
        & MLP   & \cmark & \xmark & \xmark   & \mstd{60.5}{1.2} \\
        & RF    & \cmark & \xmark & \xmark   & \mstd{61.8}{1.8} \\
        \hline
        \multirow{2}{*}{\centering \TNN} 
        & LSTM  & \cmark & \cmark & \xmark   & \mstd{62.8}{0.9} \\
        & MHA   & \cmark & \cmark & \xmark   & \mstd{62.4}{1.3} \\
        \hline
        \multirow{2}{*}{\centering \ReNN} 
        & GCN   & \cmark & \xmark & \cmark   & \mstd{64.6}{0.6} \\
        & GAT   & \cmark & \xmark & \cmark   & \mstd{65.2}{0.8} \\
        \hline
        \multirow{2}{*}{\centering \TReNN} 
        & MHA+GCN & \cmark & \cmark & \cmark  & \mstd{67.3}{0.6} \\
        & MHA+GAT & \cmark & \cmark & \cmark  & \mstd{67.1}{0.7} \\
        \hline
        \multirow{1}{*}{\centering \TEReNN} &
        \textbf{Time-Expanded GCN} & \cmark & \cmark & \cmark  
        & $\mathbf{70.1{\scriptstyle \pm 1.0}}$ \\
        \bottomrule
    \end{tabular}%
    }
    \caption{Average macro-F1 over ten seeds using a leave-one-team-out evaluation protocol. The proposed Time-Expanded GCN achieves the best performance (bold).}
    \label{tab:placeholder}
\end{table}


\subsection{Q2: Counterfactual}

\begin{figure}[t!]
    \centering
    \includegraphics[width=0.48\linewidth]{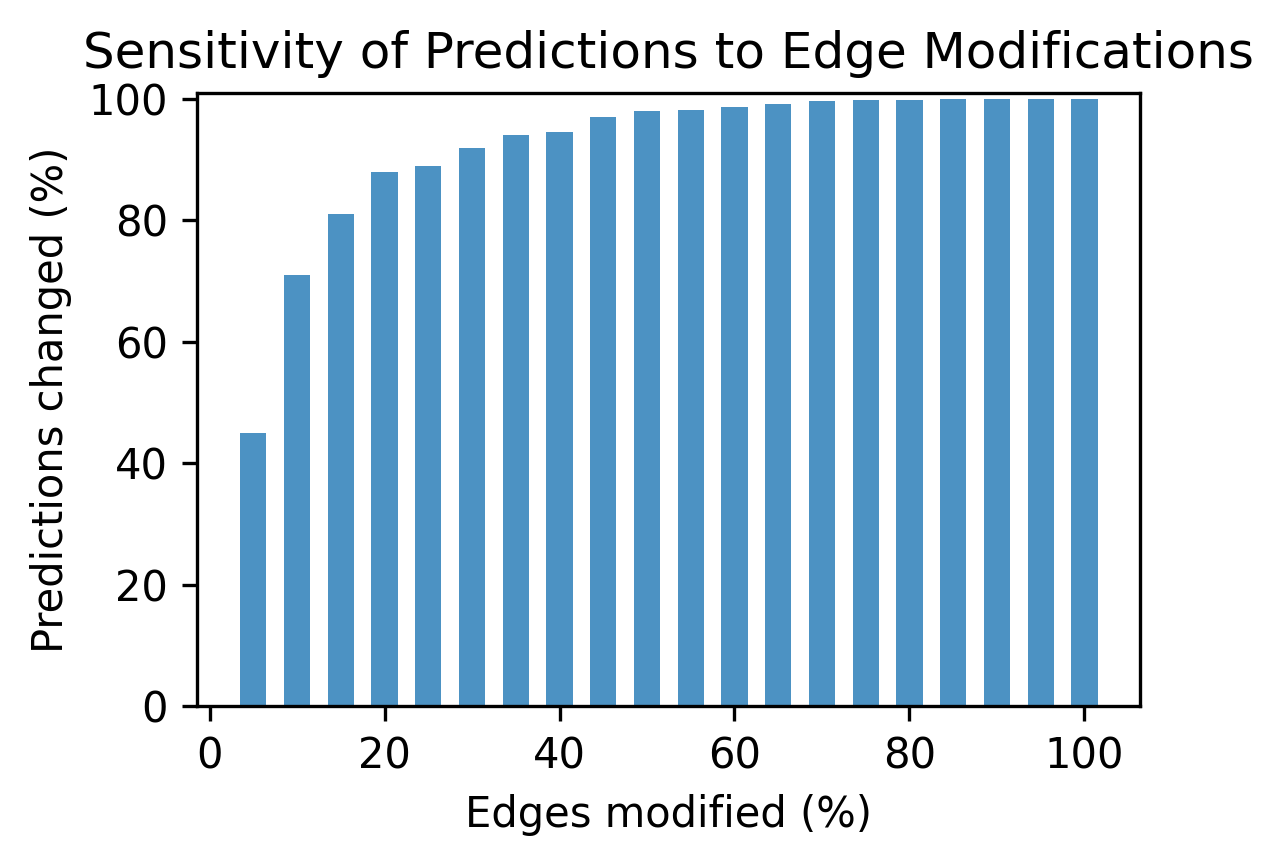}
    \hfill
    \includegraphics[width=0.48\linewidth]{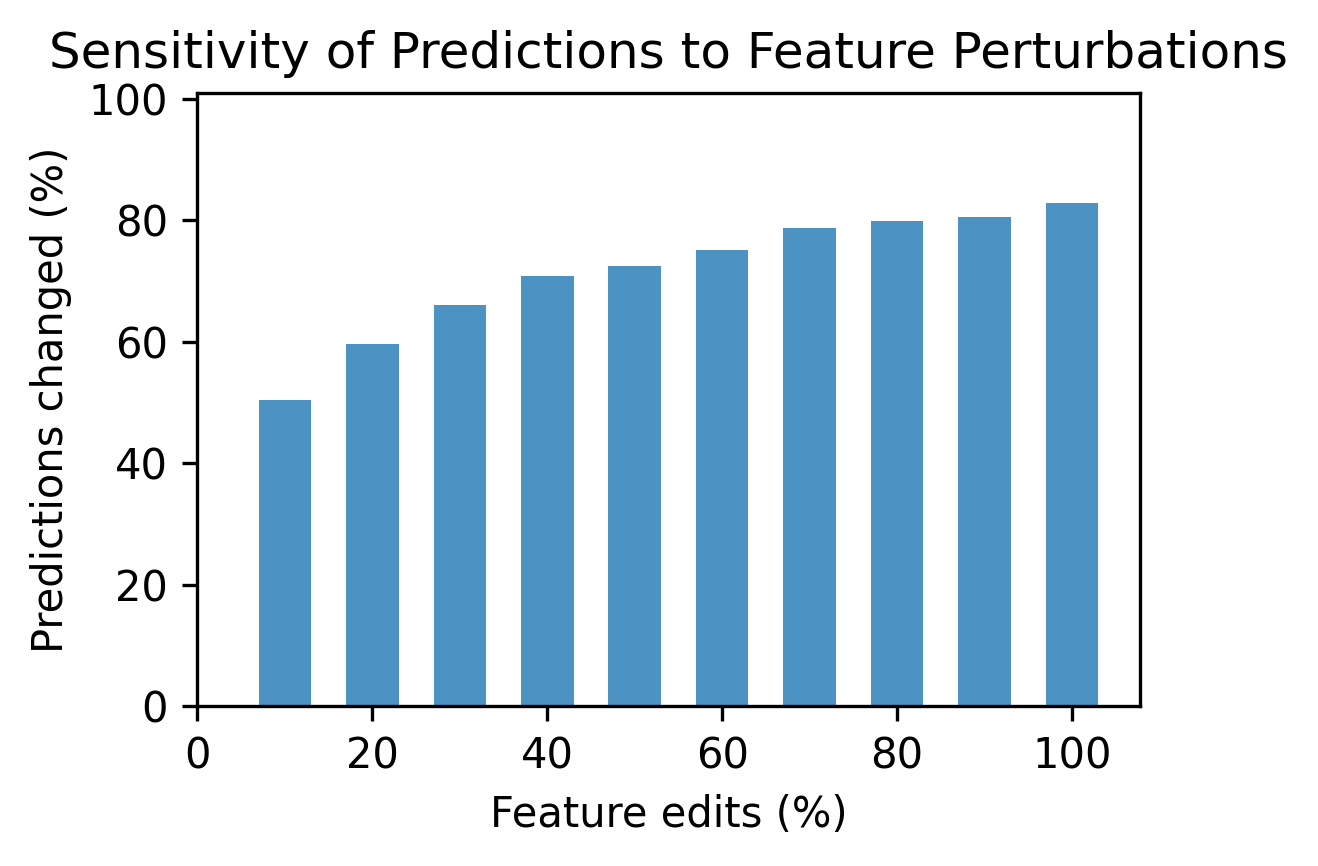}
    
    \caption{
    Sensitivity analysis of model predictions when converting a slow surgical procedure into a medium-duration one, or a medium-duration procedure into a fast one. The left panel reports the sensitivity of the predictions to modifications of interaction edges, while the right panel illustrates the sensitivity to changes in team members’ behavioral classes.}
    \label{fig:sensitivity_comparison}
\end{figure}

The \TEReNN architecture supports two different types of counterfactual explanations: topological-level and feature-level explanations.
In the case of topological counterfactuals (see Sec.~\ref{sec:counterfactual_xai}), we used a combinatorial algorithm and assessed it by determining the minimum number of interactions that must be removed to 
increase the speed of a surgical procedure (from slow to medium and from medium to fast), as shown in Fig.~\ref{fig:sensitivity_comparison}(left). 
Visual inspection of the dataset revealed that slowdowns often occur when surgeons acting as democratic leaders engage in conversations with team members who are not directly involved in a specific part of the surgery. Conversely, surgeons acting as autocratic leaders, who focus on the task and avoid unnecessary conversations, tend to speed up procedures.
For feature counterfactuals (see Sec.~\ref{sec:counterfactual_xai}), we employed a greedy algorithm on paralinguistic features, assessing the minimum distance from the original behavioral class required to switch from a slow procedure to a mid-duration procedure, and from a mid-duration to a fast procedure, as illustrated in Fig.~\ref{fig:sensitivity_comparison}(right). The sensitivity in this case is smaller than for topological counterfactuals, since class switches are performed without a gradient-based procedure and silence is not considered as a possible behavioral class. Additionally, once the behavioral class of a team member is changed, it is approximated to the average of the new class over the entire time window. Notably, modifying only 10\% of the behavioral class over a time window can produce approximately a 50\% increase in predicted performance, suggesting that even small adjustments can have a substantial impact on the model's predicted performance. Values are measured cumulatively. Slowdowns are often associated with the calm-leader behavioral class for the leader.





\section{Conclusion and Limitations}

We presented \TEReNN for modeling team dynamics and performance in the OR. Our approach outperforms competitors in predictive accuracy while leveraging the time-expanded graphs structure to provide actionable and interpretable counterfactual explanations for surgical teams. While the results are promising, we acknowledge that the operative time is only an approximation of team performance, and the practical utility of counterfactual explanations requires further validation by domain experts and can be further extended to other modalities.

\section*{Acknowledgments}
Funded by the European Union. Views and opinions expressed are however those of the author(s) only and do not necessarily reflect those of the European Union or the European Health and Digital Executive Agency (HaDEA). Neither the European Union nor the granting authority can be held responsible for them. Grant Agreement no. 101120763 - TANGO. 
GV and VMDL acknowledge the support of the MUR PNRR project FAIR - Future AI Research (PE00000013) funded by the NextGenerationEU.

\bibliographystyle{plain}
\bibliography{biblio}

\end{document}